\pdfoutput=1

\documentclass[10pt,twocolumn,letterpaper]{article}

\usepackage[pagenumbers]{iccv} 

\definecolor{iccvblue}{rgb}{0.21,0.49,0.74}
\usepackage[pagebackref,breaklinks,colorlinks,allcolors=iccvblue]{hyperref}
\usepackage{bm}
\usepackage{dsfont}
\usepackage{colortbl}
\usepackage{xcolor}
\usepackage{algorithm}
\usepackage{algorithmic}
\usepackage{hyperref}
\usepackage{url}
\usepackage{subcaption}
\usepackage{graphicx}
\usepackage{amsmath}
\usepackage{amssymb}
\usepackage{adjustbox}
\usepackage{makecell}
\usepackage{multirow}
\usepackage{bbding}
\usepackage{color, colortbl}
\definecolor{citecolor}{HTML}{2980b9}
\definecolor{linkcolor}{HTML}{c0392b}
\usepackage{enumitem}
\usepackage{amsmath}
\usepackage{algorithm}
\usepackage{algorithmic}
\usepackage{amsthm}
\newtheorem{theorem}{Theorem}
\usepackage[utf8]{inputenc} 
\usepackage[T1]{fontenc}    
\usepackage{hyperref}       
\usepackage{url}            
\usepackage{booktabs}       
\usepackage{amsfonts}       
\usepackage{nicefrac}       
\usepackage{microtype}      
\usepackage{xcolor}         
\usepackage{amsmath}
\usepackage{wrapfig} 
\usepackage{circledsteps}
\usepackage{dsfont}
\usepackage{pifont}
\usepackage{xcolor}
\usepackage{float}

\def\paperID{1273} 
\def\confName{ICCV}
\def\confYear{2025}

\title{Singular Value Fine-tuning for Few-Shot Class-Incremental Learning}

\author{
  Zhiwu Wang\textmd{\textsuperscript{1}}\footnotemark[1]\textmd{,} ~Yichen Wu\textmd{\textsuperscript{2}}\footnotemark[1]\textmd{,} ~Renzhen Wang\textmd{\textsuperscript{1}}\footnotemark[2]\textmd{,} ~Haokun Lin\textmd{\textsuperscript{2,3}}\textmd{,} ~Quanziang Wang\textmd{\textsuperscript{1}}\textmd{,} \\
  ~Qian Zhao\textmd{\textsuperscript{1}}\footnotemark[2], ~Deyu Meng\textmd{\textsuperscript{{1}}} \\
  \textsuperscript{1} Xi'an Jiaotong University, 
  \textsuperscript{2} City University of Hong Kong, \\ \textsuperscript{3} Institute of Automation, Chinese Academy of Sciences\\
  {\small
  \texttt{\{wangzhiwu7\ wuyichen.am97\}@gmail.com}\quad \texttt{rzwang@xjtu.edu.cn}\quad
  \texttt{haokun.lin@cripac.ia.ac.cn}
  }\\
  {\small
  \texttt{\{quanziangwang\ timmy.zhaoqian\}@gmail.com}\quad \texttt{dymeng@mail.xjtu.edu.cn}
  }
}

\begin{document}

\maketitle
\def\thefootnote{*}\footnotetext{These authors contribute equally to this work}
\def\thefootnote{$\dagger$}\footnotetext{Corresponding authors}

\begin{abstract}
Class-Incremental Learning (CIL) aims to prevent catastrophic forgetting of previously learned classes while sequentially incorporating new ones. The more challenging Few-shot CIL (FSCIL) setting further complicates this by providing only a limited number of samples for each new class, increasing the risk of overfitting in addition to standard CIL challenges.  While catastrophic forgetting has been extensively studied, overfitting in FSCIL, especially with large foundation models, has received less attention. To fill this gap, we propose the \textbf{S}ingular \textbf{V}alue \textbf{F}ine-tuning for FS\textbf{C}I\textbf{L} (\textbf{SVFCL}) and compared it with existing approaches for adapting foundation models to FSCIL, which primarily build on Parameter Efficient Fine-Tuning (PEFT) methods like prompt tuning and Low-Rank Adaptation (LoRA). Specifically, SVFCL applies singular value decomposition to the foundation model weights, keeping the singular vectors fixed while fine-tuning the singular values for each task, and then merging them. This simple yet effective approach not only alleviates the forgetting problem but also mitigates overfitting more effectively while significantly reducing trainable parameters. Extensive experiments on four benchmark datasets, along with visualizations and ablation studies, validate the effectiveness of SVFCL. The code will be made available.
\end{abstract}    
\section{Introduction}
\label{sec:intro}
Given the unprecedented increase in computational resources and data availability, the ability to continuously learn from new data while retaining previously acquired knowledge is crucial for developing truly intelligent systems. Class-Incremental Learning (CIL) \cite{zhou2023deepCILsurvey, rebuffi2017icarl, li2017LwF, kirkpatrick2017EWC} seeks to address this challenge by introducing new classes sequentially, requiring the model to adapt to changes in data distribution while preserving knowledge of previously learned classes. However, the practical implementation of CIL is limited by the difficulty of acquiring sufficiently large datasets for new classes, which significantly constrains its application in real-world scenarios, particularly in dynamic environments.

Few-Shot Class-Incremental Learning (FSCIL) \cite{tao2020fewTOPIC, tian2024FSCILsurvey, zhang2021fewCEC, peng2022fewALICE, zhou2022forwardFACT, yang2023neuralcollapse} focuses on the incremental learning scenario where newly introduced classes have only a limited number of samples. Unlike traditional CIL, FSCIL begins by training on a task with a sufficiently large dataset to establish prior knowledge (\textit{i.e.} the base session), followed by subsequent tasks that include only a small number of samples (\textit{i.e.} new sessions). This scenario highlights the model's ability to sequentially learn new classes from limited samples while retaining previously learned knowledge, thereby enhancing its practical applicability in real-world contexts. Conversely, the practical setting of learning incrementally with limited samples also introduces two key problems \textit{forgetting} and \textit{overfitting} in the demanding FSCIL scenario:
\begin{itemize}
    \item (\textit{Forgetting}.) During the sequential training, the model will experience a significant drop in performance on previously learned tasks after training on new tasks.
    \item (\textit{Overfitting}.) Due to the limited data of the new tasks, the model may overfit to these examples, which negatively impacts its performance on these tasks.
\end{itemize}

For the challenging \textit{forgetting} problem, Shi \etal \cite{shi2021FSCILflatminima} found that learning in the base session (\ie, the beginning large dataset) is more critical to final performance than adaptation to new sessions with few-shot examples. Consequently, most studies---especially those using shallow model architectures like ResNet-18---train the backbone exclusively on the base session while fine-tuning the classification head during subsequent few-shot sessions~\cite{zhou2022forwardFACT, peng2022fewALICE, wang2023fewTEEN, ahmed2024orco, yang2023neuralcollapse, akyurek2021subspaceregularizersforFSCIL, hersche2022constrainedFSCIL}. However, the capacity of these methods is limited by the backbone model, as recent studies have shown that larger foundation models are more effective at transferring knowledge and mitigating forgetting \cite{zhou2023deepCILsurvey}. Therefore, recent studies try to improve FSCIL performance by incorporating large pre-trained models, such as Vision Transformer (ViT). For example, ASP \cite{liu2024fewASP} incorporates specifically designed trainable prompts into ViT to enhance model adaptability across both base and novel sessions.
Furthermore, CLIP \cite{radford2021CLIP} has also inspired a series of multimodal-based works \cite{d2023multimodalPEFSCIL,park2024privilege,xu2024CA-CLIP,li2025ckpdfscil}, thereby providing semantic cues for visual classification tasks.
For instance, CKPD-FSCIL \cite{li2025ckpdfscil} adapts the CLIP-ViT backbone to foster plasticity incorporating LoRA \cite{hu2021lora} modules as its redundant-capacity components while freezing the knowledge-sensitive components for stability.
Most of these methods follow the CL with foundation model paradigm, adapting Prompt-tuning and LoRA to FSCIL in an effort to mitigate forgetting.

However, existing studies with foundation models primarily address forgetting while overlooking overfitting, which is particularly critical in FSCIL due to the limited examples in new sessions. Although Zhu \etal \cite{zhu2024enhanced} has examined overfitting in FSCIL, it focused on small, shallow models rather than the large foundation models commonly used today. To bridge this gap, we first investigate the overfitting issue in Prompt-tuning and LoRA methods under the FSCIL setting. We then propose a simple yet effective approach, \textbf{S}ingular \textbf{V}alue \textbf{F}ine-tuning for FS\textbf{C}I\textbf{L} (\textbf{SVFCL}), which significantly reduces parameter size while effectively mitigating overfitting, leading to improved overall performance. To sum up, our main contributions can be summarized as follows:
\begin{itemize}
    \item We investigate the overfitting issue in current Prompt-Tuning and LoRA methods with foundation models, showing that while they effectively mitigate forgetting, they still suffer from severe overfitting.
    \item We propose SVFCL, which first applies Singular Value Decomposition and then incrementally fine-tunes the singular values to adapt to new tasks.
    \item We conduct extensive experiments on three commonly used datasets, where SVFCL achieves state-of-the-art results in FSCIL, with comprehensive ablation studies further demonstrating its effectiveness and robustness.
\end{itemize}
\section{Related Works}
\label{sec:related}

\textbf{Class-Incremental Learning (CIL)} can be divided into four mainstream categories. \textit{Regularization-based} methods~\cite{kirkpatrick2017EWC, chaudhry2018riemannianwalk, zenke2017SI, aljundi2018MAS} typically introduce a penalty term to constrain updates on crucial model weights, thereby preserving previously acquired knowledge. In contrast, \textit{Replay-based} methods~\cite{rebuffi2017icarl, buzzega2020darkexperiencereplay, de2021continualprototypeevolution, chaudhry2018riemannianwalk, bang2021rainbowmemory, wu2024meta, wang2024dual} leverage a limited memory buffer to store past exemplars and develop strategies for utilizing these samples in an efficient and effective manner. \textit{Gradient projection} methods~\cite{saha2021gradient_projection_memory, zhao2023rethinking_gradient_projection, kong2022balancing, wang2023cba, wu2024mitigating} aim to adjust gradient directions to mitigate forgetting. Meanwhile, \textit{Architecture-based} methods~\cite{yan2021dynamicallyexpandablerepresentation, li2019learntogrow, rusu2016progressiveneuralnetworks} either learn masks to preserve important weights or expand the model, enabling other weights or newly introduced neurons to adapt to new tasks. Even though these methods help mitigate forgetting, researchers are increasingly exploring large foundation models for better knowledge retention and transfer in class-incremental learning \cite{wang2022l2p, wang2022dualprompt, smith2023coda, liang2024inflora}. For instance,
L2P~\cite{wang2022l2p} first introduces the concept of prompt pool to dynamically select task-specific prompts during incremental learning. Building on this, DualPrompt~\cite{wang2022dualprompt} incorporates task-invariant prompts to capture shared knowledge across diverse tasks. CodaP~\cite{smith2023coda} further refines this approach by enhancing both the training and selection strategies in an end-to-end framework. In addition to the prompt-tuning strategy, Low-Rank Adaptation (LoRA) \cite{hu2021lora} has also gained widespread attention in CIL \cite{liang2024inflora, wu2025s-lora}. For example, InfLoRA \cite{liang2024inflora} fine-tunes the task-specific LoRA modules in the subspace to eliminate the interference of the new task on the old tasks, thus mitigating forgetting.

\vspace{1mm}
\noindent \textbf{Parameter-Efficient-Fine-Tuning (PEFT)} aims to adapt large pre-trained models to diverse downstream tasks efficiently by incorporating extra lightweight learnable parameters while freezing the pre-trained backbone, thereby circumventing the computational and storage costs associated with direct fine-tuning of large models. With the rapid development of large pre-trained models, a growing number of PEFT methods have emerged, such as Prefix-tuning \cite{li2021prefix}, Prompt-tuning \cite{lester2021powerprompt,jia2022visualprompttuning,piao2024federated}, Adapter-tuning \cite{houlsby2019PETtransferlearningfornlpAdapter,yang2024continual}, Scale and Shift tuning~\cite{lian2022scalingandshifting}, and Low-Rank Adaption~\cite{hu2021lora,sun2022singularvalueFT,wu2025s-lora}. 

\vspace{1mm}
\noindent \textbf{Few-Shot Class-Incremental Learning (FSCIL)} was first introduced by Tao \etal~\cite{tao2020fewTOPIC} and aims to incrementally learn new few-shot tasks (i.e. new sessions) after an initial base training phase with abundant samples (i.e., the base session). Based on their focus on the base session or new sessions, current FSCIL methods can be classified into two categories: base-focus~\cite{zhang2021fewCEC, zhou2022forwardFACT, wang2023fewTEEN, yang2023neuralcollapse, akyurek2021subspaceregularizersforFSCIL, hersche2022constrainedFSCIL, ahmed2024orco, peng2022fewALICE} and incremental-focus~\cite{kim2024mics, tao2020fewTOPIC}. Recently, researchers have increasingly focused on how to leverage foundation models and parameter-efficient tuning techniques to adapt to FSCIL.
For instance, Liu \etal~\cite{liu2024fewASP} propose the Attention-aware Self-adaptive Prompt (ASP) framework with ViT as its network backbone. Park \etal~\cite{park2024privilege} propose PriViLege, which combines prompting functions and knowledge distillation. D'Alessandro \etal~\cite{d2023multimodalPEFSCIL} propose CPE-CLIP, a CLIP-based framework that utilizes prompting techniques to generalize pre-trained knowledge to incremental sessions. CKPD-FSCIL \cite{li2025ckpdfscil} incorporate LoRA \cite{hu2021lora} modules as redundant-capacity components to adapt CLIP-ViT to foster plasticity while freezing knowledge-sensitive components for stability.
However, most of these methods primarily focus on alleviating forgetting while neglecting the critical overfitting problem in FSCIL. In contrast, the proposed SVFCL demonstrates excellent performance in mitigating overfitting while achieving strong overall results.

\begin{figure*}[t]
    \centering
    \includegraphics[width=0.7\linewidth]{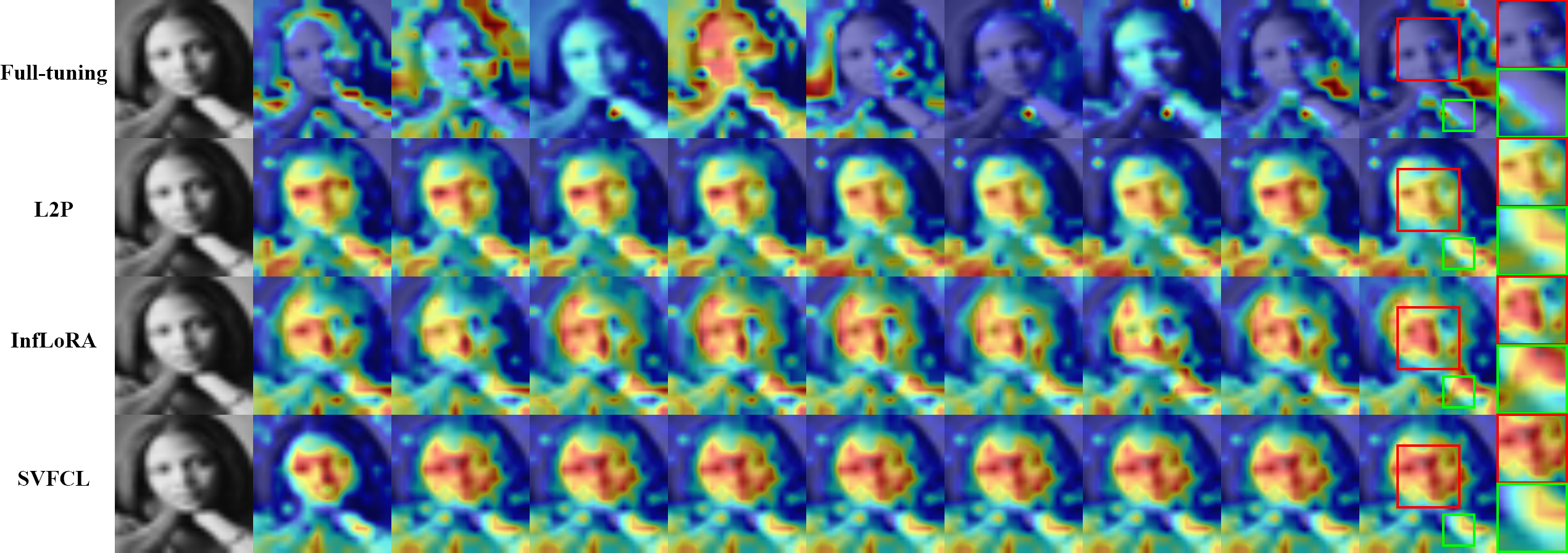}
    \caption{
    Illustration of attention maps on one sampled CIFAR-100 image across four methods during FSCIL. Our proposed approach can effectively focus on critical features while maintaining robustness. In contrast, the full fine-tuning strategy struggles to capture key attention patterns, whereas the other two representative methods, InfLoRA \cite{liang2024inflora} and L2P \cite{wang2022l2p}, tend to emphasize background features partially.
    }
    \label{fig:main_attention}
\end{figure*}
\section{Preliminary}
\label{sec:pre}

\noindent \textbf{Problem Formulation.} Suppose there is a sequence of downstream tasks denoted by $\{\mathcal T^0, \mathcal T^2, \cdots, \mathcal T^{M}\}$, where the training set for the $t$-th task is given by $\mathcal{D}^t=\{(\bm{x}_i^t,y_i^t)\}_{i=1}^{|\mathcal D_t|}$. Here, $\bm{x}_i^t$ represents the input image and $y_i^t\in C^t$ denotes the corresponding label. Following the disjoint FSCIL settings proposed by Tao \etal \cite{tao2020fewTOPIC}, we adopt the non-overlapping label spaces $C^t$ for different tasks, \textit{i.e.},  $C^i\cap C^j=\emptyset$ for all \(i\neq j\), where $i,j \in \{0,1,...,M\}$.
FSCIL begins with a base session $\mathcal D^0$, which contains a sufficient amount of data. In contrast, each subsequent new session $\mathcal{D}^t (t>0)$ includes only a limited number of training samples for new classes in $C^t$. For instance, in an $N$-way $K$-shot FSCIL setting, each incremental session $\mathcal D^t (t>0)$ consists of $N$ classes, with each class containing $K$ samples. The objective of FSCIL is to ensure that the model performs well on both the fully sampled base session and the subsequent new sessions with limited examples. Formally, this can be expressed as
\begin{equation}
\begin{aligned}
    &\min_{\bm \phi}\mathbb E_{(\bm{x} ,y)\in \{\mathcal D^i\}_{i=0}^M}\mathcal L(F_{\bm \phi}(\bm{x} ),y) \\
    = &\min_{\bm \theta, \bm \omega}\mathbb E_{(\bm{x} ,y)\in\{\mathcal D^i\}_{i=0}^M}\mathcal L(g_{\bm w}\circ f_{\bm \theta}(\bm{x} ),y),
\end{aligned}
\label{min_risk_target}
\end{equation}
where $\mathcal L(\cdot,\cdot)$ denotes the Cross-entropy loss function, 
the model $F_{\bm \phi}(\cdot)$ can be further decomposed into an image encoder \(f_{\bm \theta}\) parameterized by \(\bm \theta\) and a classifier \(g_{\bm \omega}\) parameterized by $\bm \omega$, such that $F_{\bm \phi}=g_{\bm \omega}\circ f_{\bm \theta}$.

\vspace{1mm}
\noindent \textbf{Singular Value Decomposition (SVD)} is a powerful tool for uncovering intrinsic properties and structures in datasets. Specifically, SVD expresses the fixed model weights matrix $\mathbf{W}\in\mathbb R^{m\times n}$ as:
\begin{equation}
    \mathbf{W} =  \mathbf{U}\mathbf{\Sigma}\mathbf{V}^\intercal
    \label{eq:svd}
\end{equation}
where \(\mathbf{U}\in \mathbb{R}^{m\times m}\) is an orthogonal matrix whose columns are the left singular vectors, \(\mathbf{\Sigma}\in \mathbb{R}^{m\times n}\) is a diagonal matrix containing the non-negative singular values, and \(\mathbf{V}^\intercal \in \mathbb{R}^{n\times n}\) is an orthogonal matrix whose columns are the right singular vectors of \(\mathbf{W}\).

\vspace{1mm}
\noindent \textbf{Prompt-tuning} learns a set of trainable prompt embeddings $\bm P$ that can be inserted at various locations in the model, including the input layer and intermediate layers. Given an input $\bm x$, prompt-tuning updates its representation as follows:
\begin{equation*}
    h = F(\left[\bm x; \bm P\right];\bm \phi).
\end{equation*}
By integrating prompts into the network, prompt-tuning enables task adaptation while keeping weights unchanged.

\vspace{1mm}
\noindent \textbf{LoRA} addresses parameter efficiency by decomposing the weight update $\Delta \mathbf{W}$ into a low-rank factorization:
\begin{equation*}
    \Delta \mathbf{W} = \mathbf{AB},
\end{equation*}
where $\mathbf{A}\in{\mathbb{R}^{m\times r_1}}$ and $\mathbf{B}\in{\mathbb{R}^{r_1\times n}}$ with $r_1\ll \min\{m,n\}$. By adding the updated low-rank matrices $\mathbf{A}$ and $\mathbf{B}$ to the fixed $\mathbf{W}$, LoRA facilitates adaptation to new tasks.
\section{Method}
\label{sec:method}
In this section, we will first analyze the overfitting problem associated with current popular PEFT techniques, such as Prompt-tuning and LoRA, when adapting foundation models for FSCIL. Next, we introduce the proposed SVFCL method and present both empirical and theoretical analyses demonstrating how it alleviates the overfitting issue in FSCIL.

\begin{table}[t]
    \centering
    \resizebox{0.85\linewidth}{!}{
    \begin{tabular}{ccc}
    \toprule
    Method & Number (M) & Percentage (\%) \\
    \midrule
    Full-tuning & 86.6 & 100 \\
    L2P~\cite{wang2022l2p} & 0.485 & 0.56 \\
    InfLoRA~\cite{liang2024inflora} & 0.261 & 0.27 \\
    \textbf{SVFCL (Ours)} & \textbf{0.095} & \textbf{0.11} \\
    \bottomrule
    \end{tabular}
    }
    \caption{Number (Million) and percentage (\%) of learnable parameters required for tuning across various approaches using the  ViT-B/16 backbone.}
    \label{tab:params}
\end{table}

\begin{figure}[h]
    \centering
    \includegraphics[width=.7\linewidth]{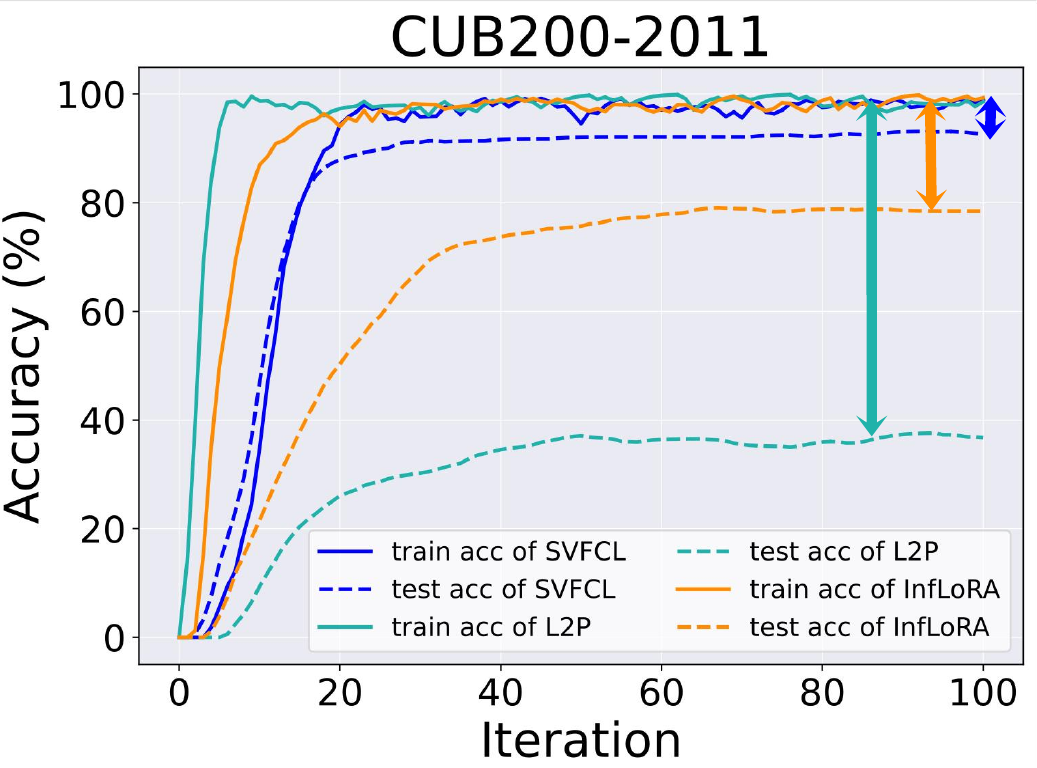}
    \caption{Illustration of the accuracy curve on both the training and validation datasets across three approaches during the first incremental few-shot session. InfLoRA \cite{liang2024inflora} and L2P \cite{wang2022l2p} face a serious risk of overfitting, while the proposed SVFCL demonstrates a strong ability to mitigate overfitting.}
    \label{ab:overfitting}
\end{figure}
\begin{figure*}[h]
    \centering
    \includegraphics[width=0.8\linewidth]{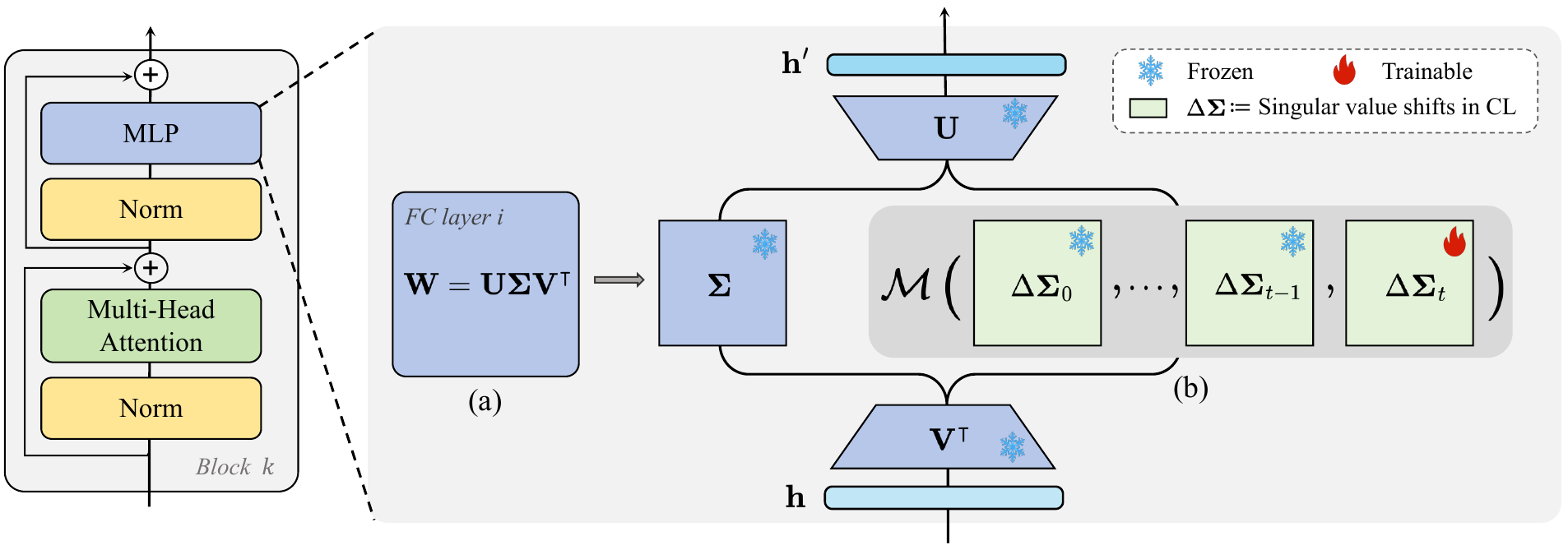}
    \caption{The framework of the proposed SVFCL algorithm. (a) We first perform singular value decomposition on the pre-trained weights $\mathbf W$ and obtain the fixed singular matrices $\mathbf{U}$ and $\mathbf{V}$. (b) We incrementally fine-tune the singular values on the current training task and get the singular value shift $\Delta \mathbf{\Sigma}_t$. The notation $\mathcal{M}(\cdot)$ denote the merging function to fuse all learned singular shifts as shown in Eqn. (\ref{eq:merging}).}
    \label{fig:pinepline}
\end{figure*}
\subsection{Overfitting in Prompt-Tuning and LoRA}
To analyze the overfitting issue in Prompt-tuning and LoRA, we first introduce the representative methods: the Prompt-tuning method L2P~\cite{wang2022l2p} and the LoRA-based method InfLoRA~\cite{liang2024inflora}. In \cref{tab:params}, we compare their trainable parameters, showing that both methods significantly reduce the number of trainable parameters compared to full fine-tuning. Additionally, we visualize the attention maps of the corresponding models on a CIFAR-100 image in Fig.~\ref{fig:main_attention} and plot the accuracy on both the training and validation datasets on CUB200-2011
in \cref{ab:overfitting}. While these methods help mitigate overfitting by reducing trainable parameters and encouraging focus on discriminative features, the visualizations highlight persistent issues. The attention maps in \cref{fig:main_attention} reveal that a significant portion of the model's attention remains on the background, rather than on the more informative features. Furthermore, the noticeable gap between the accuracy on the training and validation sets in \cref{ab:overfitting} serves as clear evidence of overfitting. These two visualizations together highlight the inability of both prompt-tuning and LoRA to generalize effectively, despite the reduction in parameter sizes.

The overfitting observed may stem from the inherent characteristics of the methods themselves. In Prompt-tuning, prompts implicitly influence specific layers rather than directly updating model parameters. While this reduces trainable parameters, it still increases the risk of fitting noise, especially in FSCIL's few-shot setting with limited new task samples~\cite{ma2023understanding}. Similarly, LoRA-based methods require learning new directions (\ie $\mathbf{A}$ and $\textbf{B}$ matrices) for each task, introducing additional flexibility that aids adaptation. However, this flexibility also increases the risk of overfitting, particularly when dealing with scarce data in new tasks. 

\begin{algorithm}[t]
    \caption{Training Algorithm of the Proposed SVFCL}
    \label{alg}
    \begin{algorithmic}[1]
        \STATE \textbf{Input:} A sequence of tasks $\{\mathcal D^0, \mathcal D^1\cdots,\mathcal D^M\}$ with label spaces \(\{C^i\}_{i=0}^M\), pre-trained image encoder \(f_{\bm\theta}\), random initialized NCM classifier \(g_{\bm \omega}\), pre-trained weight matrix \(\mathbf W\) contained in \(\bm\theta\) for adaptation.\\
        
        \textcolor{lightgray}{\# Before FSCIL}
        \STATE Apply SVD: \(\mathbf W=\mathbf U\mathbf \Sigma\mathbf V^\intercal\)
        
        \textcolor{lightgray}{\# During FSCIL} 
        \FOR{current task \(\mathcal D^t\)}
            \STATE Add a new adapter: \(\Delta\mathbf\Sigma_t\)
            \STATE Merge all seen adapters by Eqn. (\ref{eq:merging})
            \STATE Freeze all other parameters except \(\Delta\mathbf\Sigma_t\) and \(\bm\omega\)
            \STATE Do forward process on \(\mathcal D^t\)
            \STATE Update $\Delta \mathbf\Sigma_t$ and \(\bm\omega\) by Eqn. (\ref{min_risk_target})
            \FOR{new class \(c\in C^t\)}
                \STATE Calculate prototype \(\bm{p}_c\)
                \STATE Update the NCM classifier: \(\bm\omega[c]\leftarrow \bm{p}_c\)
            \ENDFOR
        \ENDFOR
        \RETURN \(\bm\theta,\bm\omega\)
    \end{algorithmic}
\end{algorithm}

\subsection{The Proposed SVFCL Algorithm}
\label{subsec: svfcl}
To address the overfitting issues in Prompt-tuning and LoRA-based methods, we propose the \textbf{S}ingular \textbf{V}alue \textbf{F}ine-tuning FS\textbf{C}I\textbf{L} method (\textbf{SVFCL}), which aims to enhance adaptation in new sessions with limited examples in FSCIL while maintaining the adaptability of foundation models to new tasks.  
Specifically, for the foundation model weights $\mathbf W \in \mathbb{R}^{m \times n}$, we perform singular value decomposition as $\mathbf W=\mathbf U\mathbf \Sigma\mathbf V^\intercal$ and fix the singular matrices $\mathbf{U}\in \mathbb{R}^{m\times m}$ and $\mathbf{V}^\intercal\in \mathbb{R}^{n\times n}$ throughout the training process. As shown in \cref{fig:pinepline}, once the proposed SVFCL completes training for the current $t$-th session, the updated weights $\mathbf{W}_t$ can be expressed
\begin{equation}
\label{eq:merging}
\begin{split}
    \mathbf{W}_t= &\mathbf{W} + \Delta\mathbf{W} \\
    = &\mathbf{W} + \mathbf{U}\mathcal{M}\left(\{\Delta\bm\Sigma_i\}_{i=0}^{t-1}, \Delta \bm \Sigma_t\right)\mathbf{V}^\intercal,
\end{split}
\end{equation}
where $\mathcal{M}(\cdot)$ denotes the merging function that combines the fine-tuned singular values across sequential training sessions. $\Delta \bm \Sigma_t$ represents the singular values learned for the current task, while $\{\Delta \bm \Sigma_i\}_{i=1}^{t-1}$ refers to singular values learned for previous tasks and is kept fixed.

Note that the update $\Delta \mathbf{W}$ can also be expressed in an alternative form as the sum of several rank-one matrix products, as follows
\begin{equation}
\label{eqn:rank-1}
    \Delta \mathbf{W} = \sum_{k=1}^{r_2} \mathcal{M}( \{\sigma_i^k\}_{i=0}^t) \bm{u}_i\bm{v}_i^\intercal,
\end{equation}
where $r_2\le\min\{m,n\}$ denotes the rank of $\mathbf{W}$, $\sigma_i^k$ means the $k$-th singular value of $\Delta \bm \Sigma_i$, and $\bm u_i$, $\bm v_i$ represent the $i$-th left and right singular vectors, respectively.

As shown in Eqn.~(\ref{eqn:rank-1}), the proposed SVFCL can be viewed as the summation of multiple low-rank matrix products (\textit{i.e.}, $\bm u_i\bm v_i^\intercal$), where the low-rank matrices are fixed and only the coefficients are updated. This means that the update $\Delta \mathbf{W}$ in SVFCL remains within the span of the left and right singular vectors, and the weight space basis does not change during the update. In contrast, LoRA allows updates outside of $\mathbf{U}$ and $\mathbf{V}$, effectively modifying the basis itself. While this provides greater flexibility for adapting to new tasks, it also increases the risk of overfitting, especially in FSCIL settings where new sessions have only a few examples.

\subsection{Theoretical Analysis}
Considering the superior performance of the LoRA-based method in mitigating overfitting compared to prompt-tuning, as shown in ~\cref{ab:overfitting}, and its similarity to the proposed SVFCL, we provide a mathematical justification in this section. Specifically, we analyze two key perspectives to explain why SVFCL is better suited for FSCIL than LoRA.

\vspace{1mm}
\noindent
\textbf{\textit{Generability.}} LoRA-based methods~\cite{liang2024inflora,wu2025s-lora} introduces {\small $\Delta \mathbf{W}_{\mathrm{LoRA}} = \sum_{i=1}^{t-1}\mathbf{A}_i\mathbf{B}_i + \mathbf{A}_t\mathbf{B}_t$} with the two matrices {\small $\mathbf{A}_t\in{\mathbb{R}^{m\times r_1}}$ and $\mathbf{B}_t\in{\mathbb{R}^{r_1\times n}}$} are trainable, requiring {\small $O(r_1(m + n))$} parameters, while the proposed SVFCL only adjust the singular values {\small $\Delta \mathbf{W} = \sum_{k=1}^{r_2} \mathcal{M}( \{\sigma_i^k\}_{i=1}^t) \bm{u}_i\bm{v}_i^\intercal$}, requiring only {\small $O(r_2)$} parameters. As shown in ~\cref{tab:params}, the proposed SVFCL has much less trainable parameters compared to InfLoRA~\cite{liang2024inflora}.

According to the VC dimension theory~\cite{vapnik2015uniform}, models with fewer parameters generalize better when trained on limited data. This is further validated by the visualizations in \cref{fig:main_attention} and \cref{ab:overfitting}, as well as the experimental results in \cref{sec:exp}.

\begin{theorem}[\textit{\textbf{Optimization Stability.}}]
    Let {\small $\mathbf{W}\in\mathbb{R}^{m\times n}$} be the pre-trained weight matrix, and {\small $\mathbf{W}^*\!\in\! \mathbb{R}^{d\times k}$} be the optimal weight matrix for both the current task and previous ones. Without loss of generality, assume the rank parameters $r_1=r_2=r$. Suppose {\small $\Delta \mathbf{W}^*=\mathbf{W}^*\!-\!\mathbf{W}$} lies within the subspace spanned by the eigenvectors of {\small $\mathbf{W}$}. Then, we can have
    \begin{equation*}
    \begin{split}
     \|\Delta \mathbf{W}_{\rm{SVF}}\|_F &=\| \mathbf{U}\mathcal{M}\left(\{\Delta\bm\Sigma_i\}_{i=0}^{t-1}, \Delta \bm \Sigma_t\right)\mathbf{V}^\intercal\|_F\\
     & = \sqrt{\sum_{k=1}^r\mathcal{M}( \{\left[\sigma_i^k\right]^2\}_{i=0}^t  )}\\
     & \le  \|\sum_{i=0}^t\mathbf{A}_i\mathbf{B}_i\|_F =  \|\Delta \mathbf{W}_{\mathrm{LoRA}}\|_F.
    \end{split}
    \end{equation*}
\end{theorem}
It can be seen that the proposed SVFCL’s updates are energy-concentrated on the principal components, minimizing unnecessary perturbations to the pre-trained weights. In contrast, LoRA’s updates risk altering less relevant directions, increasing the likelihood of overfitting in incremental learning. We provide the detailed proof using Eckart-Young-Mirsky theorem \cite{eckart1936approximation} in supplementary materials.

\begin{table*}[t]
\begin{center}
\resizebox{.9\linewidth}{!}{
\begin{tabular}{lccccccccccccccc}
\toprule
\multirow{2}{*}{\textbf{Methods}} & \multicolumn{5}{c}{\textbf{miniImageNet} (\%)} & \multirow{2}{*}{\(A_{avg}\)\(\uparrow\)} & \multirow{2}{*}{\(PD\)\(\downarrow\)} &
\multicolumn{6}{c}{\textbf{CUB200-2011} (\%)} & \multirow{2}{*}{\(A_{avg}\)\(\uparrow\)} & \multirow{2}{*}{\(PD\)\(\downarrow\)} \\
\cmidrule{2-6}\cmidrule{9-14}
& 0 & 2 & 4 & 6 & 8 & & & 0 & 2 & 4 & 6 & 8 & 10 &\\
\midrule
\color{gray}{\textit{Conventional CL \& FSCIL}} \\
iCaRL \cite{rebuffi2017icarl} & 97.0 & 94.0 & 91.5 & 87.2 & 85.6 & 90.9 & 11.4 
& 88.4 & 80.2 & 73.1 & 67.6 & 65.7 & 63.3 & 72.7 & 25.1 \\
FOSTER \cite{wang2022foster} & 96.9 & 89.0 & 88.2 & 85.5 & 83.9 & 88.2 & 13.0 
& 88.1 & 81.6 & 76.6 & 73.6 & 72.0 & 72.3 & 77.2 & 15.8 \\
CEC \cite{zhang2021fewCEC} & 87.4 & 84.0 & 83.1 & 80.7 & 80.7 & 83.0 & 6.7 
& 84.8 & 81.4 & 79.3 & 77.4 & 77.2 & 76.8 & 79.1 & 8.0 \\
FACT \cite{zhou2022forwardFACT} & 72.6 & 66.4 & 60.6 & 54.3 & 50.5 & 60.7 & 22.1
& 87.3 & 82.1 & 78.4 & 75.4 & 74.4 & 73.9 & 78.2 & 13.4 \\
TEEN \cite{wang2023fewTEEN} & 97.3 & 95.2 & 94.4 & 92.4 & 92.1 & 94.3 & 5.2 
& 89.1 & 84.7 & 81.3 & 78.8 & 77.6 & 77.4 & \underline{81.1} & 11.7 \\
\midrule
\color{gray}{\textit{CL \& FSCIL w/ FD models}} \\
L2P$^\ddagger$ \cite{wang2022l2p} & 97.2 & 84.4 & 74.0 & 65.8 & 59.4 & 75.8 & 37.8
& 85.5 & 73.5 & 65.6 & 58.7 & 52.3 & 49.4 & 63.8 & 36.1 \\
DualP$^\ddagger$ \cite{wang2022dualprompt} & 97.5 & 83.6 & 73.2 & 65.0 & 58.6 & 75.2 & 38.9 
& 85.8 & 74.0 & 65.5 & 58.0 & 51.7 & 49.5 & 63.7 & 36.3 \\
CodaP$^\ddagger$ \cite{smith2023coda} & \textbf{98.0} & 88.7 & 82.5 & 74.6 & 70.1 & 82.5 & 28.0 
& \textbf{89.2} & 78.5 & 70.4 & 63.4 & 56.9 & 55.7 & 68.5 & 33.5 \\
InfLoRA$^\ddagger$ \cite{liang2024inflora} & 97.4 & 90.1 & 78.8 & 69.7 & 62.7 & 80.0 & 34.7
& 89.0 & 79.7 & 66.9 & 58.3 & 51.3 & 46.0 & 65.2 & 43.0 \\
CPE-CLIP$^*$ \cite{d2023multimodalPEFSCIL} & 90.2 & 87.4 & 86.5 & 83.4 & 82.8 & 86.1 & 7.4
& 81.6 & 76.7 & 71.5 & 67.7 & 65.1 & 64.6 & 70.8 & 17.0 \\
PriViLege$^\ddagger$ \cite{park2024privilege} & 96.7 & 95.6 & 95.5 & 94.3 & 94.1 & \underline{95.3} & \underline{2.6}
& 82.2 & 80.4 & 77.8 & 75.7 & 75.2 & 75.1 & 77.5 & \underline{7.1} \\
CKPD-FSCIL$^*$ \cite{li2025ckpdfscil} & 96.2 & 92.8 & 91.1 & 87.0 & 86.2 & 90.1 & 9.9 
& 87.1 & 82.3 & 76.8 & 77.5 & 76.5 & 75.6 & 79.0 & 11.5 \\
\midrule
\rowcolor{orange!10}
\textbf{SVFCL (Ours)}$^\dagger$ & 97.6 & \textbf{96.5} & \textbf{96.4} & \textbf{95.5} & \textbf{95.3} & \textbf{96.3} & \textbf{2.3}
& 87.1 & \textbf{85.0} & \textbf{83.7} & \textbf{81.9} & \textbf{82.2} & \textbf{82.6} & \textbf{83.5} & \textbf{4.5} \\
\bottomrule
\end{tabular}
}
\end{center}
\caption{Performance comparison on miniImageNet and ImageNet-R across three evaluation metrics: Top-1 accuracy \(A_t\) for each task, average accuracy \(A_{avg}\), and performance dropping (\(PD\)). The notation
$^{\dagger}$ indicates the backbone pre-trained on ImageNet-21K and fine-tuned on ImageNet-1K, $^\ddagger$ indicates teh backbone pre-trained on ImageNet-21K, and \(^*\) indicates the CLIP backbone.
}
\label{tab:main_results_mini_cub}
\end{table*}

\section{Experiments}
\label{sec:exp}

\subsection{Experimental Settings}

\textbf{Datasets.}
Following prior works \cite{zhang2021fewCEC, zhou2022forwardFACT, wang2023fewTEEN}, we conduct experiments on three datasets, CUB200-2011~\cite{wah2011caltech_used_birds_200_2011_dataset_CUB200}, miniImageNet~\cite{ravi2016optimization_as_a_model_for_fewshot_learning_MiniImageNet} and ImageNet-R \cite{hendrycks2021many_imagenetr} to validate the effectiveness and the robustness of our approach. For CUB200-2011 and ImageNet-R, we set the base session with 100 classes and adopt the 10-way 5-shot setting, \ie 10 classes per session and 5 samples per class, for the subsequent 10 sessions, while for miniImageNet, we allocate 60 classes for the base session and adopt the 5-way 5-shot setting for the subsequent 8 sessions. Please refer to the supplementary materials for a detailed introduction to these datasets.

\vspace{1mm}
\noindent\textbf{Evaluation Protocol.} Following D'Alessandro \etal \cite{d2023multimodalPEFSCIL}, we assess the performance using two evaluation metrics, including Average Accuracy (\(A_{avg}\)) and Performance Dropping (\(PD\)). Given the Top-1 accuracy \(A_t\) for the \(t\)-th task, the two metrics are defined as follows: (1) \(A_{avg}\), computed as \(A_{avg} = \frac{1}{M} \sum_{t=1}^M A_t\), which measures the mean performance across all the seen tasks after training the $M$-th task; and (2) \(PD\), defined as \(PD = A_0 - A_M\), which quantifies the extent of forgetting between the first and the last task.

\vspace{1mm}
\noindent\textbf{Baselines.} We mainly compare our proposed method with recent state-of-the-art approaches, which can be broadly categorized into 
two groups: (1) Conventional CL and FSCIL methods: iCaRL \cite{rebuffi2017icarl}, FOSTER \cite{wang2022foster}, CEC \cite{zhang2021fewCEC}, FACT \cite{zhou2022forwardFACT}, TEEN \cite{wang2023fewTEEN}. These methods employ various strategies to mitigate forgetting but primarily focus on shallow networks, such as ResNet-18; 
(2) CL and FSCIL methods with FD models: L2P~\cite{wang2022l2p}, DualPrompt~\cite{wang2022dualprompt}, CodaP~\cite{smith2023coda} incorporate the Prompt-tuning strategy within CL, aiming to select task-specific prompts during training and testing to mitigate forgetting. InfLoRA~\cite{liang2024inflora} incrementally updates low-rank matrices and employs orthogonal regularization to reduce forgetting. Meanwhile, CPE-CLIP~\cite{d2023multimodalPEFSCIL}, PriViLege~\cite{park2024privilege}, CKPD-FSCIL \cite{li2025ckpdfscil} leverage the more powerful ViT or CLIP model and multimodal information to address forgetting in the FSCIL setting. Note that for a fair comparison, L2P, DualPrompt, CodaP, InfLoRA, and our proposed SVFCL all use the ViT-B/16 backbone pre-trained on ImageNet-21K and fine-tuned on ImageNet-1K. We replace the backbone for conventional CL and FSCIL methods with the same one we use to reproduce the results. Despite CPE-CLIP, PriViLege, and CKPD-FSCIL utilizing a significantly more powerful backbone, SVFCL still achieves superior performance.

\vspace{1mm}
\noindent\textbf{Implementation Details.} Following~\cite{wang2022l2p, wang2022dualprompt, smith2023coda,liang2024inflora},
we choose the ViT-B/16-1K ~\cite{alexey2020ViT_an_image_is_worth_16x16_words_transformers_for_image_recognition_at_scale}, pre-trained on ImageNet-21K and fine-tuned on ImageNet-1K, as the backbone. Inspired by Zhang \etal \cite{zhang2023adalora}, who highlighted that fine-tuning the Feed-Forward Network is more effective than fine-tuning the self-attention module, we exclusively fine-tune the pre-trained weight matrices in the MLP across all 12 blocks of the ViT backbone. And we conduct ablation studies to explore different choices of adapted blocks. For the merging function $\mathcal{M}(\cdot)$ in Eqn.~(\ref{eqn:rank-1}), we simply employ the summation strategy.
Additionally, we replace the original linear classifier with the Nearest Class Mean (NCM) classifier after the backbone, following prior works \cite{zhang2021fewCEC, zhou2022forwardFACT, wang2023fewTEEN}. For all datasets, the input images are resized to 224\(\times\)224. Across all three datasets, we train the base session for five epochs and each few-shot session for two epochs, ensuring consistent evaluation of the model’s performance. The training optimizer is set to Adam~\cite{Adam} with a learning rate of 5e-4 throughout the training phase. Moreover, our experiments are all implemented using PyTorch~\cite{paszke2019pytorch} on an NVIDIA GeForce RTX 4090. 

\begin{figure*}[h]
    \centering
    \begin{subfigure}{.3\linewidth}
        \centering
        \includegraphics[width=.95\linewidth]{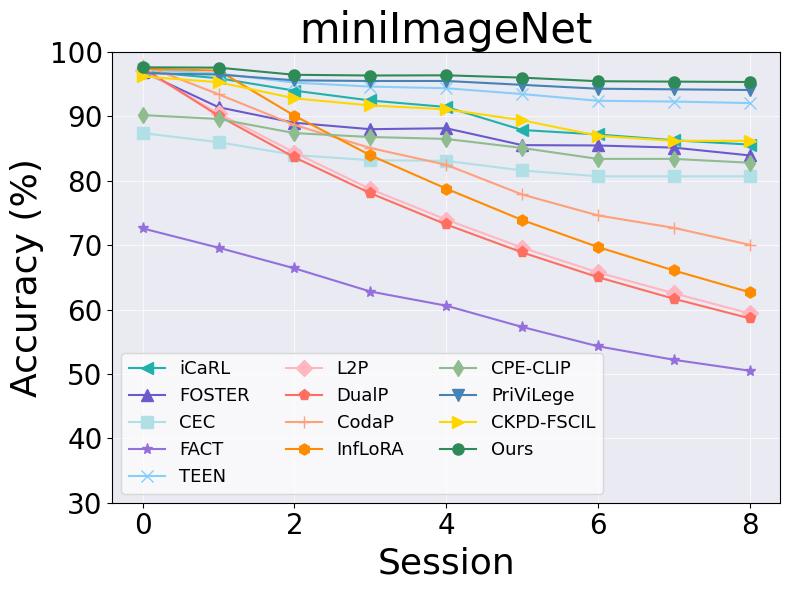}
    \end{subfigure}
    \begin{subfigure}{.3\linewidth}
        \centering
        \includegraphics[width=.95\linewidth]{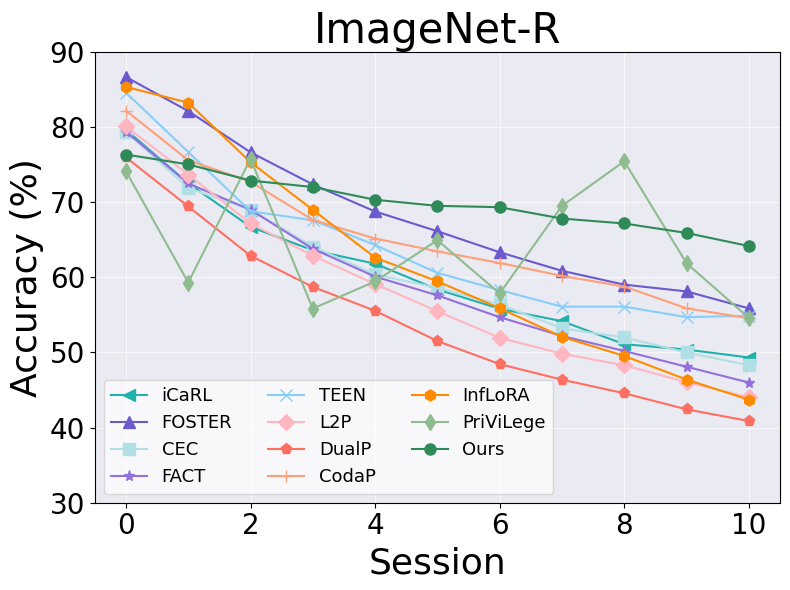}
    \end{subfigure}
    \begin{subfigure}{.3\linewidth}
        \centering
        \includegraphics[width=.95\linewidth]{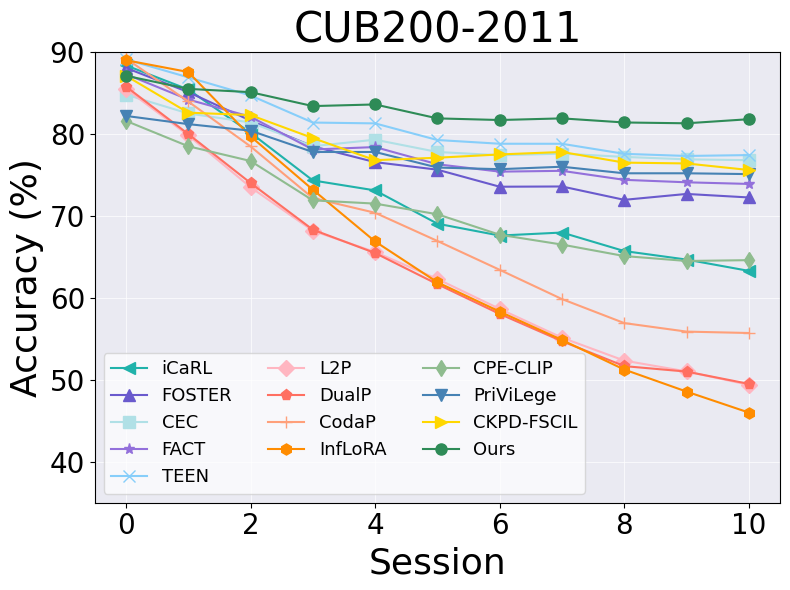}
    \end{subfigure}
    \caption{Illustration of Top-1 accuracy curves during sequential training on the miniImageNet, ImageNet-R, and CUB200-2011 datasets.}
    \label{fig:comparison_all}
\end{figure*}

\subsection{Main Results}
In this section, we report the comparison results of SVFCL with the aforementioned baselines on the three datasets as shown in \cref{tab:main_results_mini_cub} and \cref{tab:main_results_inr}. The detailed accuracy of each session on the three datasets is shown in~\cref{fig:comparison_all}.

Quantificationally, SVFCL achieves the highest Top-1 average accuracy of 96.3\%, 70.1\%, and 83.5\%, with the lowest performance drops of 2.3\%, and 12.2\%, and 4.5\%, on miniImageNet, ImageNet-R, and CUB200-2011, respectively. 

In detail, on the miniImageNet dataset, SVFCL achieves a notable 1.0\% performance improvement in Top-1 accuracy over the second-best method, PriViLege, which even uses the stronger backbone pre-trained on ImageNet-21K.
Similarly, in evaluations on CUB200-2011, SVFCL also exhibits significant performance gains, outperforming the leading conventional FSCIL method, TEEN, by 2.4\% and surpassing the CLIP-based CKPD-FSCIL, by 4.5\%. On ImageNet-R, SVFCL exceeds other baselines, delivering an average accuracy 1.9\% higher than the second-best method, FOSTER.
~These results consistently demonstrate the robust capacity of our method in learning incremental tasks and mitigating catastrophic forgetting. Besides, we can observe that prompt-based CL methods sometimes fail to outperform conventional FSCIL methods such as CEC and FACT, despite being specifically designed to leverage strong foundation models. This limitation arises from their susceptibility to overfitting. In contrast, our method fine-tunes the model across all sessions, as demonstrated in Fig.~\ref{ab:overfitting}, effectively mitigating overfitting while maintaining strong performance with minimal forgetting.
It is worth noting that ImageNet-R, serving as an out-of-distribution dataset in relation to the ImageNet-21K dataset, is a suitable dataset for evaluating the pre-trained ViT model.

To demonstrate the effectiveness of our method, we also compare the magnitude of learnable parameters across various approaches, including the widely adopted Prompt-tuning (L2P) and LoRA (InfLoRA) in our analysis. As shown in \cref{tab:params}, our proposed SVFCL achieves the lowest number of learnable parameters. For instance, SVFCL requires fewer than half the learnable parameters of InfLoRA. 
This substantial reduction plays a key role in mitigating overfitting in few-shot tasks. Moreover, despite this decrease in the number of learnable parameters, the representational capacity of our method remains robustly preserved, ensuring its effectiveness across diverse scenarios, as evidenced by our experimental results shown in \cref{tab:main_results_mini_cub} and \cref{tab:main_results_inr}.

\begin{table}[t]
    \centering
    \resizebox{0.6\linewidth}{!}{
    \begin{tabular}{cc}
        \toprule
        Blocks & Average Accuracy (\%) \\ \midrule
        0-6   & 83.46 \\
        3-8   & 83.12 \\
        9-11  & 82.80 \\ \bottomrule
    \end{tabular}
    }
    \caption{Comparison results of employing SVFCL in 6 different blocks in ViT on CUB200-2011.}
    \label{teb:ab_top6blocks_cub}
\end{table}

\begin{figure*}[h]
    \centering
    \begin{minipage}[t]{0.32\textwidth}
        \centering
        \includegraphics[width=\linewidth]{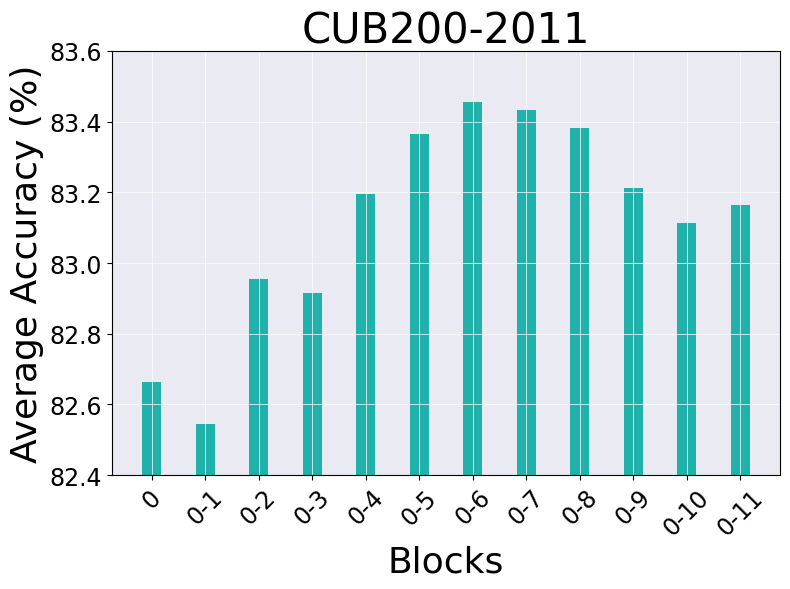} 
        \caption{Ablation study of fine-tuning different blocks within ViT.}
        \label{fig:ab_block_cub}
    \end{minipage}
    \hfill 
    \begin{minipage}[t]{0.65\textwidth}
        \centering
        \includegraphics[width=\linewidth]{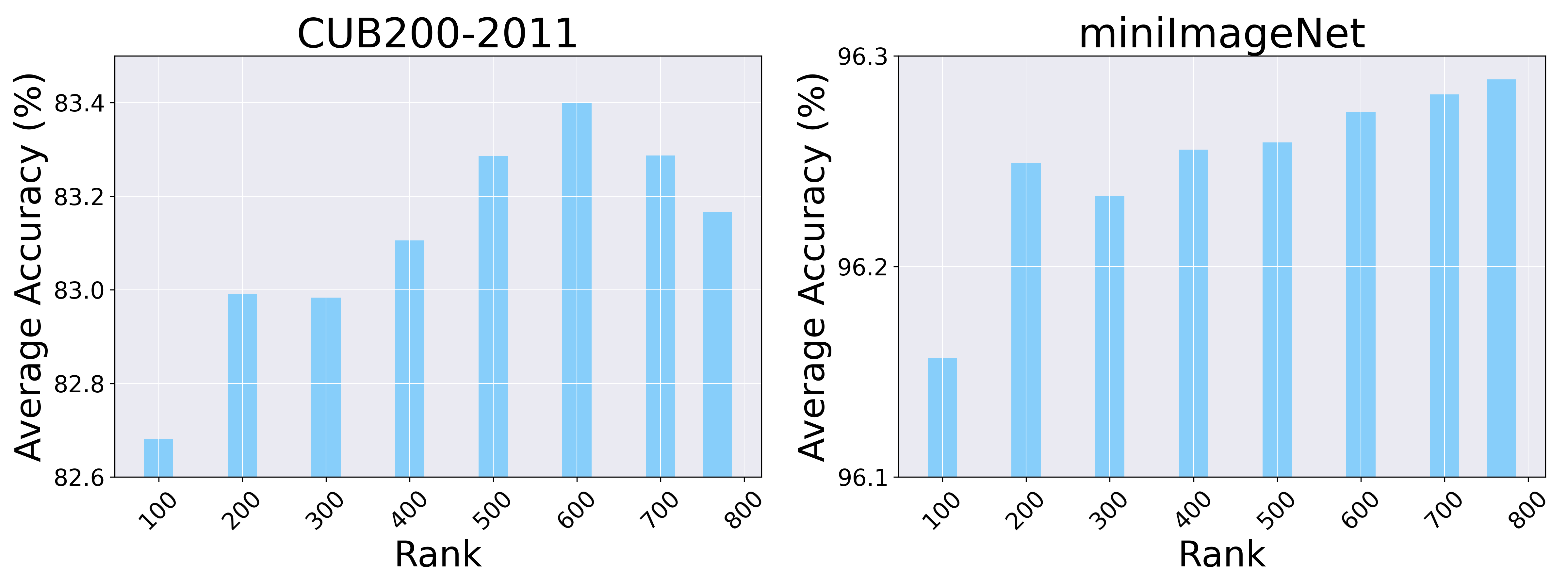}
        \caption{Ablation study of truncated SVD fine-tuning in SVFCL on the CUB200-2011 and miniImageNet datasets with different ranks.}
        \label{fig:ab_rank}
    \end{minipage}
\end{figure*}

\subsection{Discussion}

\begin{table}[t]
\begin{center}
\resizebox{.95\linewidth}{!}{
\begin{tabular}{lcccccccc}
\toprule
\multirow{2}{*}{\textbf{Methods}} & \multicolumn{6}{c}{\textbf{Top-1 accuracy on ImageNet-R (\%)}} & \multirow{2}{*}{\(A_{avg}\)\(\uparrow\)} & \multirow{2}{*}{\(PD\)\(\downarrow\)} \\
\cmidrule{2-7}
& 0 & 2 & 4 & 6 & 8 & 10  \\
\midrule
\color{gray}{\textit{Conventional CL \& FSCIL}} \\
iCaRL \cite{rebuffi2017icarl} & 79.7 & 66.7 & 61.8 & 55.8 & 51.1 & 49.3 & 60.3 & 30.3 \\
FOSTER \cite{wang2022foster} & \textbf{86.7} & 76.7 & 68.8 & 63.3 & 59.0 & 55.9 & \underline{68.2} & 30.8 \\
CEC \cite{zhang2021fewCEC} & 79.4 & 69.0 & 60.4 & 56.4 & 52.0 & 48.3 & 60.3 & 31.1 \\
FACT \cite{zhou2022forwardFACT} & 79.4 & 69.0 & 60.1 & 54.7 & 50.2 & 46.0 & 59.4 & 33.4 \\
TEEN \cite{wang2023fewTEEN} & 84.6 & 68.8 & 64.3 & 58.3 & 56.1 & 54.9 & 63.9 & 29.7 \\
\midrule
\color{gray}{\textit{CL \& FSCIL w/ FD}} \\
L2P$^{\dagger}$ \cite{wang2022l2p} & 80.1 & 67.2 & 59.1 & 52.0 & 48.3 & 44.0 & 58.1 & 36.1 \\
DualP$^{\dagger}$ \cite{wang2022dualprompt} & 76.0 & 62.9 & 55.6 & 48.5 & 44.6 & 40.9 & 54.3 & 35.1 \\
CodaP$^{\dagger}$ \cite{smith2023coda} & 82.2 & 72.7 & 65.2 & 61.9 & 58.8 & 54.6 & 65.3 & 27.7 \\
InfLoRA$^{\dagger}$ \cite{liang2024inflora} & 85.4 & 75.3 & 62.6 & 55.9 & 49.5 & 43.7 & 62.1 & 41.6 \\
PriViLege$^{\ddagger}$ \cite{park2024privilege} & 74.1 & \textbf{75.8} & 59.5 & 57.9 & 75.5 & 54.6 & 64.4 & \underline{19.5} \\
\midrule
\rowcolor{orange!10}
\textbf{SVFCL (Ours)}$^\dagger$ & 76.4 & 72.9 & \textbf{70.3} & \textbf{69.3} & \textbf{67.2} & \textbf{64.2} & \textbf{70.1} & \textbf{12.2} \\
\bottomrule
\end{tabular}
}
\end{center}
\caption{Performance comparison on the ImageNet-R dataset across three evaluation metrics: Top-1 accuracy \(A_t\) for each task, average accuracy \(A_{avg}\), and performance dropping (\(PD\)). 
}
\label{tab:main_results_inr}
\end{table}

To gain a deeper understanding of the effectiveness of the proposed SVFCL model, we conduct extensive discussions and answer the following questions.

\vspace{1mm}
\noindent\textbf{Does data leakage interfere with the evaluation of our method?} Our proposed method SVFCL leverages the ViT backbone pre-trained on ImageNet, raising the question that the performance gains might be attributed to data leakage, particularly if the distribution of downstream datasets overlaps significantly with that of ImageNet. To investigate this concern, we conducted additional experiments on ImageNet-R, a dataset designed to exhibit a substantial distributional shift from ImageNet, often considered out-of-distribution (OOD) due to its inclusion of artistic renditions, sketches, and other stylized variations of ImageNet classes. The performance of SVFCL on ImageNet-R shown in \cref{tab:main_results_inr} demonstrates that our method maintains robust performance, achieving the best Top-1 average accuracy, 70.1\%. This indicates that our method’s performance gains do not result entirely from data leakage or overfitting to ImageNet features. Its strong generalization to ImageNet-R highlights its ability to effectively handle diverse, unseen distributions.

\medskip
\noindent\textbf{Where to employ SVFCL within ViT?}
As aforementioned in our implementation details, we can apply SVFCL across different blocks in our ViT backbone. Intuitively, applying SVFCL to more blocks may enhance plasticity for new tasks but also increase the number of learnable parameters, raising the risk of overfitting. To strike a balance between efficiency and performance, we conduct an ablation study on the CUB200-2011 dataset, progressively applying SVFCL to an increasing number of blocks. The results are illustrated in \cref{fig:ab_block_cub}, indicating that using SVFCL in blocks 0-6 yields the optimal performance.
Additionally, we compare the impact of applying SVFCL to the same number of blocks at different positions in ViT, as shown in \cref{teb:ab_top6blocks_cub}. The results reveal that fine-tuning the first 6 blocks significantly outperforms fine-tuning the last 6 blocks. Based on these findings, we can insert SVFCL into the 0-6 blocks of the pre-trained ViT to gain a better performance.

\medskip
\noindent\textbf{Can learnable parameters be further reduced in SVFCL?}
In the proposed SVFCL, we utilize full SVD to reconstruct the pre-trained weights. However, the full SVD requires tuning all singular values for each few-shot session, which may be redundant because the pre-trained weights in each block are typically low rank, as validated in supplementary materials. Consequently, we replace the full SVD with a low-rank approximation.
Specifically, we choose \(r^\prime<r_2\) to reconstruct the weight matrix \(\mathbf W\) such as \(\mathbf W=\sum_{i=1}^{r^\prime}\sigma_i\bm{u}_i\bm{v}_i^\intercal\), in which we only fine-tune the top-\(r^\prime\) singular values. To demonstrate the feasibility of the low-rank approximation SVD in SVFCL, we conduct experiments on two datasets CUB200-2011 and miniImageNet, as shown in \cref{fig:ab_rank}. The results show that the performance of low-rank approximation SVD significantly impacts the performance of SVFCL, and excessively coarse approximations, such as setting the rank to 50 or 100, leading to performance degradation on both datasets. Within an acceptable range, we can set the rank \(r^\prime\) to 500, thereby reducing model parameters and enabling efficient fine-tuning without substantially degrading the performance of our method.

\begin{figure}[h]
    \centering
    \includegraphics[width=0.7\linewidth]{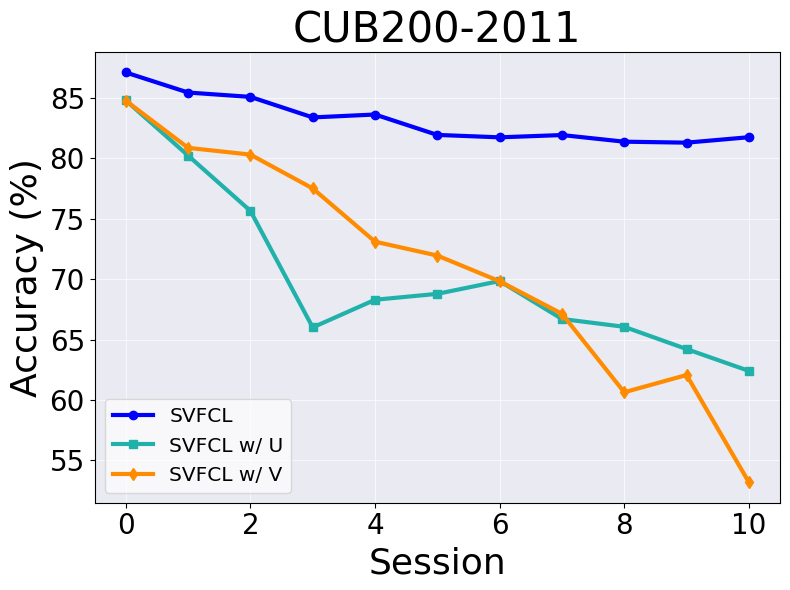}
    \caption{Ablation study of fine-tuned components: U and singular values, V and singular values, and only singular values.}
    \label{fig:ab_uv_cub}
\end{figure}

\medskip
\noindent
\textbf{Why do we freeze the singular vectors?}
The pre-trained weights can be viewed as the span of the corresponding bases, i.e., \(\bm u_1\bm v_1^\top, \cdots, \bm u_r\bm v_r^\top\), as discussed in Eqn. (\ref{eqn:rank-1}). Only fine-tuning the singular values (scales) with the frozen bases maintains the important generative knowledge of the pre-trained model, leading to a rapid adaptation when facing new tasks. We conduct additional ablation experiments to further investigate the influence of \(\mathbf U, \mathbf V\). As shown in \cref{fig:ab_uv_cub}, when additionally fine-tuning \(\mathbf U\) and \(\mathbf V\), the performance declines significantly, suggesting that preserving the original bases is crucial for retaining the model’s foundational capabilities. This observation underscores the trade-off between adaptability and stability, highlighting the effectiveness of the fine-tuning approach in our method to selectively fine-tune the most important components (singular value) while freezing other parameters.

\section{Conclusion}
\label{sec:con}
In this work, we address the challenges of catastrophic forgetting and overfitting in the Few-Shot Class-Incremental Learning (FSCIL) scenario, with a particular focus on leveraging pre-trained models. We propose a simple yet effective framework that introduces the SVFCL strategy to fine-tune the pre-trained Vision Transformer (ViT) backbone for each incremental session. SVFCL effectively mitigates the issues of overfitting and catastrophic forgetting while requiring relatively few learnable parameters and minimal computational overhead. Additionally, we provide analytical experiments and theoretical insights to demonstrate that our approach outperforms prompt-based and LoRA-based methods in resisting overfitting. Our experiments demonstrate that SVFCL delivers promising performance improvements compared to various state-of-the-art methods. Looking ahead, we plan to explore the potential of integrating SVFCL with other advanced architectures to further enhance its performance and applicability across diverse datasets.

{
    \small
    \bibliographystyle{ieeenat_fullname}
    \bibliography{main}
}

\clearpage
\setcounter{page}{1}
\maketitlesupplementary

\section*{Datasets Details}
Following prior works, we conduct experiments on three datasets CUB200-2011~\cite{wah2011caltech_used_birds_200_2011_dataset_CUB200}, miniImageNet~\cite{ravi2016optimization_as_a_model_for_fewshot_learning_MiniImageNet} and ImageNet-R \cite{hendrycks2021many_imagenetr} to evaluate the effectiveness and robustness of our method. We introduce the detailed information of the three datasets as follows:
\begin{itemize}
    \item \textbf{Caltech-UCSD Birds-200-2011 (CUB200-2011)} \cite{wah2011caltech_used_birds_200_2011_dataset_CUB200} is a fine-grained bird dataset, containing 11788 colored images in 200 classes. There are 5994 training images and 5794 testing images. We use 100 classes with all training images as the base session. The remaining 100 classes are divided into 10 incremental new sessions with 10-way 5-shot each, i.e. each new session contains 10 classes and 5 samples per class.
    \item \textbf{miniImageNet} \cite{ravi2016optimization_as_a_model_for_fewshot_learning_MiniImageNet} is a subset of the ImageNet \cite{deng2009imagenet} dataset, consisting of 60000 colored images in 100 classes with 600 images each. We use 60 classes as the base session and the remaining 40 classes are split into 8 incremental new sessions with 5-way 5-shot each, i.e. each new session contains 5 classes and 5 samples per class.
    \item \textbf{ImageNet-R}, short for ImageNet-Rendition, has various article renditions of objects from 200 ImageNet \cite{deng2009imagenet} \cite{hendrycks2021many_imagenetr} categories, with a total of 30000 images. ImageNet-R has a data distribution that is very different from ImageNet. Therefore, ImageNet-R is widely used to evaluate the robustness of the classification performance. We set the base session with 100 classes, while the remaining 100 classes are split into 10 few-shot sessions with 5 samples per class, \ie the 10-way 5-shot setting.
\end{itemize}

\section*{Proof of the Theorem}

\begin{theorem}[\textit{\textbf{Optimization Stability.}}]
    Let {\small $\mathbf{W}\in\mathbb{R}^{m\times n}$} be the pre-trained weight matrix, and {\small $\mathbf{W}^*\!\in\! \mathbb{R}^{d\times k}$} be the optimal weight matrix for both the current task and previous ones. Without loss of generality, assume the rank parameters $r_1=r_2=r$. Suppose {\small $\Delta \mathbf{W}^*=\mathbf{W}^*\!-\!\mathbf{W}$} lies within the subspace spanned by the eigenvectors of {\small $\mathbf{W}$}. Then, we can have
    \begin{equation*}
    \begin{aligned}
     \|\Delta \mathbf{W}_{\mathrm{SVF}}\|_F &=\| \mathbf{U}\mathcal{M}\left(\{\Delta\bm\Sigma_i\}_{i=0}^{t-1}, \Delta \bm \Sigma_t\right)\mathbf{V}^\intercal\|_F\\
     & = \sqrt{\sum_{k=1}^r\mathcal{M}( \{\left[\sigma_i^k\right]^2\}_{i=0}^t  )}\\
     & \le  \|\sum_{i=0}^t\mathbf{A}_i\mathbf{B}_i\|_F =  \|\Delta \mathbf{W}_{\mathrm{LoRA}}\|_F.
    \end{aligned}
    \end{equation*}
\end{theorem}
\begin{proof}
    Under the given assumption, the two methods achieve the optimal \(\Delta \mathbf W^*\) by their own adaptation strategies. We can reconstruct the problem to a matrix approximation problem, for example, that of SVFCL:
    \begin{equation}
    \begin{aligned}
        &\arg\min_{\Delta\bm\Sigma}\|\mathbf W_{\mathrm{SVF}}-\mathbf W^*\|,
    \end{aligned}
    \label{eqn:matrix_reconstruction}
    \end{equation}
    such that
    \begin{equation*}
        \begin{aligned}
            \mathbf W_{\mathrm{SVF}}&=
            \mathbf U\mathbf \Sigma\mathbf V^\intercal+\sum_{k=1}^r\mathcal M(\{\sigma_k^i\}_{i=0}^{t-1}, \sigma_k^t)\bm u_k\bm v_k^\intercal,
        \end{aligned}
    \end{equation*}
    where \(\sigma_k^i\) denotes the corresponding optimal singular values of SVFCL, \(t\) is the current task index and only \(\sigma_k^t\) for all \(k\) are trainable.
    Since LoRA fine-tune both \(\mathbf A\) and \(\mathbf B\), it introduces additional shifts to its eigenvectors.
    Using the Eckart-Young-Mirsky theorem \cite{eckart1936approximation}, we can directly obtain that the optimization problem \cref{eqn:matrix_reconstruction} gives the optimal reconstruction of \(\Delta\mathbf W^*\). Thus we have
    \begin{equation*}
        \|\Delta\mathbf W_{\mathrm{SVF}}\|_F\leq\|\Delta\mathbf W_{\mathrm{LoRA}}\|_F.
    \end{equation*}
\end{proof}

\section*{Analysis of Singular Values}
\label{appendix:rank}
\begin{figure}[h]
    \centering
    \includegraphics[width=0.8\linewidth]{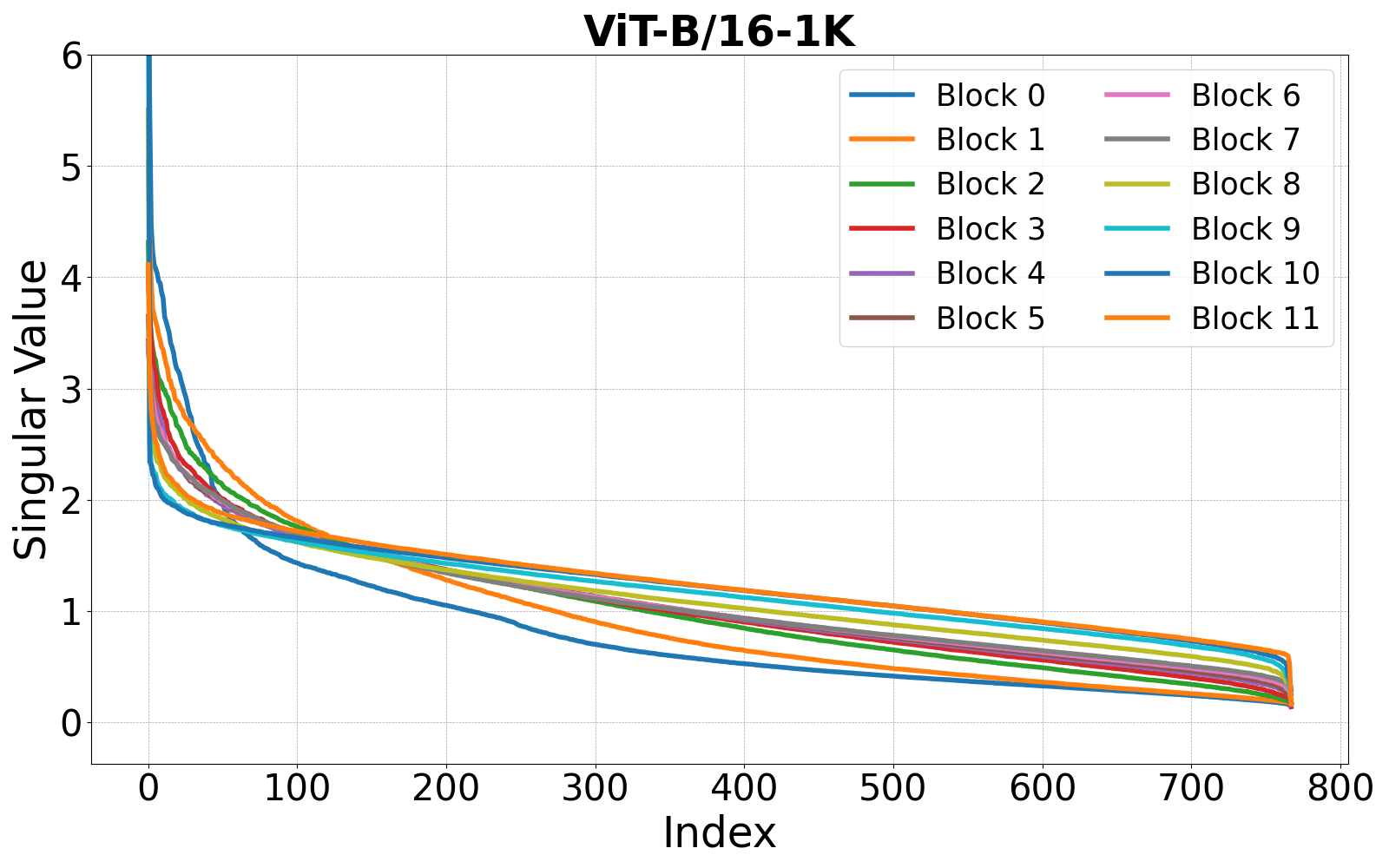}
    \caption{The singular values of the pre-trained weights of the first fully connected layers within MLP modules. The singular values exhibit a long-tailed distribution, with the majority of singular values being relatively small.}
    \label{fig:pre_sv}
\end{figure}

To examine the singular values of different layers, we visualize them across various blocks, as shown in \cref{fig:pre_sv}. The large singular values are limited, which forms the motivation for our proposed SVFCL method.

\end{document}